\documentclass{article}
\usepackage{ijcai24}

\usepackage{times}
\usepackage{soul}
\usepackage{url}
\usepackage[hidelinks]{hyperref}
\usepackage[utf8]{inputenc}
\usepackage[small]{caption}
\usepackage{graphicx}
\usepackage{amsmath}
\usepackage{amsthm}
\usepackage{booktabs}
\usepackage{algorithm}
\usepackage{algorithmic}
\usepackage[switch]{lineno}
\usepackage{gensymb}
\usepackage{bbding}

\usepackage{bm}
\usepackage{amssymb}

\usepackage{booktabs}

\usepackage{threeparttable}
 \usepackage{multirow}
 \usepackage{makecell}
 
 \usepackage{threeparttable}
 
\usepackage{hyperref}

\title{Data-Centric Evolution in Autonomous Driving: A Comprehensive Survey of Big Data System, Data Mining, and Closed-Loop Technologies}

\author{
Lincan Li$^{1,\dagger}$\and
Wei Shao$^{2}$\and
Wei Dong$^{3,\dagger}$\and
Yijun Tian$^{4,\dagger}$\and
Qiming Zhang$^{5}$\and
Kaixiang Yang$^{6}$\and
Wenjie Zhang$^{1}$\\
\affiliations
$^1$Data and Knowledge Research Group, University of New South Wales, Sydney, Australia\\
$^2$Data61, CSIRO, Clayton, Victoria, Australia\\
$^3$BYD Intelligent Driving R\&D Centre, Shenzhen, China\\
$^4$University of Notre Dame, South Bend, U.S.A\\
$^5$School of Computer Science, The University of Sydney, Sydney, Australia\\
$^6$School of Computer Science \& Engineering, South China University of Technology, Guangzhou, China\\
$^{\dagger}$ \textit{Equal Contribution}
\emails
lincan.li@unsw.edu.au,
phdweishao@gmail.com,
dong.wei13@byd.com,
yijun.tian@nd.edu,
qzha2506@uni.sydney.edu.au,
yangkx@scut.edu.cn,
wenjie.zhang@unsw.edu.au
}

\begin{document}

\maketitle

\begin{abstract}
The aspiration of the next generation's autonomous driving (AD) technology relies on the dedicated integration and interaction among intelligent perception, prediction, planning, and low-level control. There has been a huge bottleneck regarding the upper bound of autonomous driving algorithm performance, a consensus from academia and industry believes that the key to surmount the bottleneck lies in data-centric autonomous driving technology. Recent advancement in AD simulation, closed-loop model training, and AD big data engine have gained some valuable experience. However, there is a lack of systematic knowledge and deep understanding regarding how to build efficient data-centric AD technology for AD algorithm self-evolution and better AD big data accumulation. To fill in the identified research gaps, this article will closely focus on reviewing the state-of-the-art data-driven autonomous driving technologies, with an emphasis on the comprehensive taxonomy of autonomous driving datasets characterized by milestone generations, key features, data acquisition settings, etc. Furthermore, we provide a systematic review of the existing benchmark closed-loop AD big data pipelines from the industrial frontier, including the procedure of closed-loop frameworks, key technologies, and empirical studies. Finally, the future directions, potential applications, limitations and concerns are discussed to arouse efforts from both academia and industry for promoting the further development of autonomous driving. The project repository is available at: \href{https://github.com/LincanLi98/Awesome-Data-Centric-Autonomous-Driving}{https://github.com/LincanLi98/Awesome-Data-Centric-Autonomous-Driving}.
\end{abstract}

\section{Background Introduction and Motivation} \label{sec1}

The advancement of deep learning technology combined with hardware supercomputing capacity and big data collection ability, has aroused the emergence of autonomous driving algorithms, which involve a range of task modules such as detection~\cite{WuWen2023,2023querybased}, tracking~\cite{multiobject2023,Li2022CVPR}, occupancy forecasting~\cite{Khurana2023,zheng2023occworld}, trajectory prediction~\cite{Wang2022CVPR,zhangAI-TP}, motion planning~\cite{10154577,chen2023end2end}, and low-level control~\cite{kim2022end}. To achieve the ultimate goal of safe, adaptive, and efficient autonomous driving, all of the algorithms and softwares must be built upon a highly automated closed-loop data driven pipeline~\cite{2022rethinking,chen2023e2esurvey} for self-evolution and upgrading. This is the pre-requisite of the predicted large-scale commercial application of autonomous vehicle (AV) technology in year 2025~\cite{khan2022level}. Despite extensive research works have done to solely focus on designing AD algorithm itself using either open-source datasets or self-collected datasets, little efforts have been devoted to developing a systematic technology pipeline of data-centric autonomous driving, which in turn give negative impacts to the entire systems' performance improvement, resulting in the existing bottleneck of AD algorithm performance upper-bound. The above mentioned concerns motivate us to carry out a comprehensive review regarding the state-of-the-art data-centric autonomous driving technologies, including AD big data system, AD data mining, and closed-loop AD technologies. By adopting this research, we anticipate that it will enhance the industrial R\&D process and algorithm performance of autonomous driving, while also providing valuable insights to related academic studies in the same field.
\begin{figure}[t]
\centering
\includegraphics[width=0.98\columnwidth]{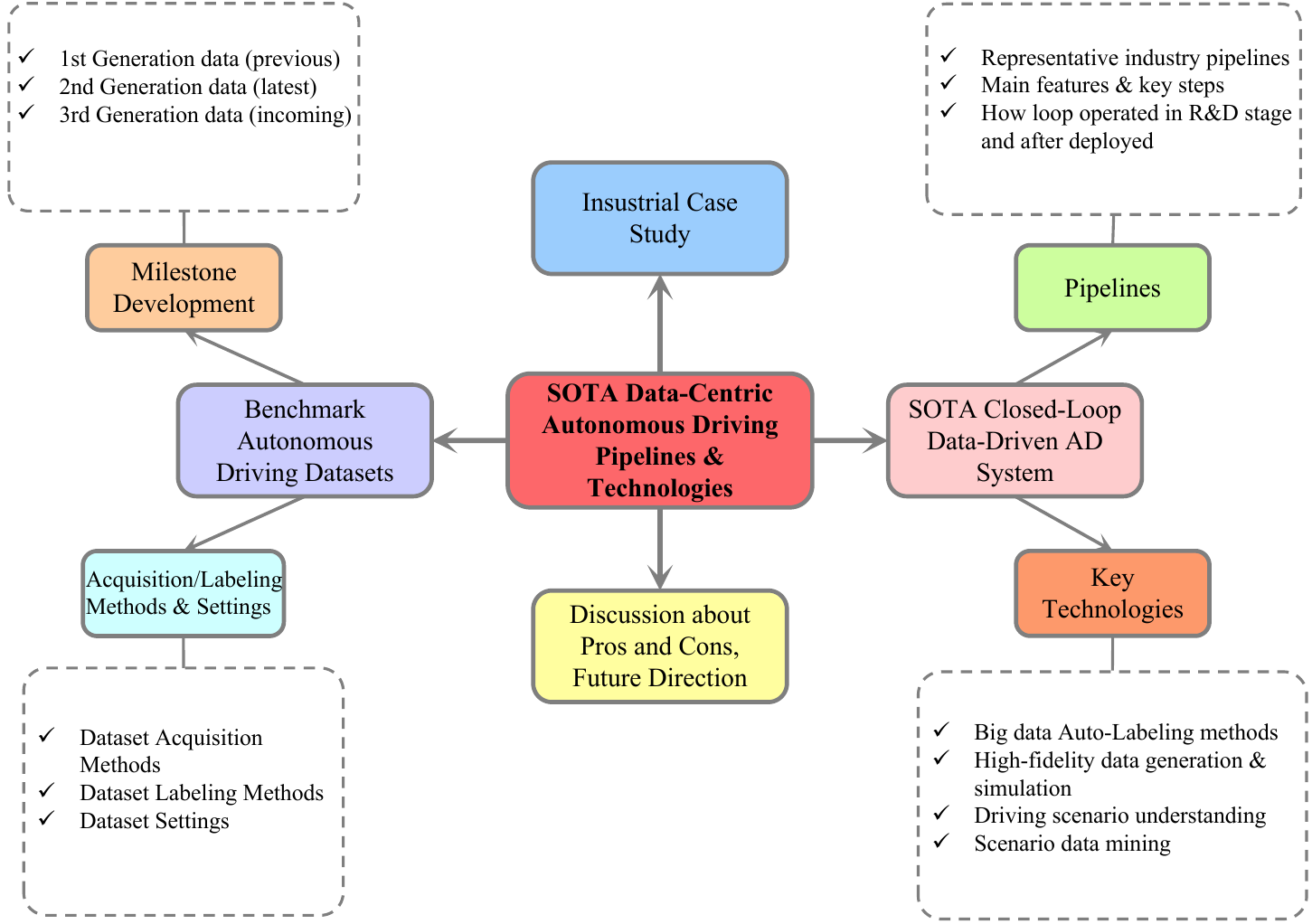}
\caption{This work provides a systematic survey of the benchmark closed-loop data-driven autonomous driving technologies from various aspects, the organization of this paper is illustrated above.}
\label{fig1}
\end{figure}

Data is the new oil in autonomous driving world. There has been a severe Long-Tail Distribution problem in real-world autonomous driving applications~\cite{long2022retrieval}. In order to fully address the long-tail distribution problem, it is estimated that we need to accumulate more than 1000 Billion driving mileage's big data~\cite{9971752}, which is far from being realized yet. Therefore, the big question arises: how to solve the long-tail distribution problem? There are three insights regarding to the answer. First, when encounter an unfamiliar condition/corner case, AD algorithms become unreliable and may make wrong decisions. However, such instances are exceedingly rare in the database, and are challenging to be acquired from real-world. Under this circumstance, AD data mining and advanced data generation technologies hold promise for addressing the problem. Second, closed-loop data-driven approaches are indispensable for advanced AD algorithm development and deployment, with an emphasize on automatically alleviating long-tail distribution problem. Third, we should re-consider the way of collecting, storing, and utilizing the massive autonomous driving data. Data should be collected across all types of essential sensors equipped on AVs, not only camera recorded driving videos. During the storage and utilization procedure, information privacy, anonymity, and security should be guaranteed.

There are a few pioneer industrial practices in data-driven autonomous driving~\cite{opML-NV,wav2023T,wav2023M}. Among them, Tesla is a representative, with its long-term developed Fleet Learning pipeline~\cite{Tesla01} and AutoPilot system. Whenever a corner case is detected by the AutoPilot, there will be a quick "snapshot" record of the case, which is comprised of all the major sensors' data during 1-minute time. After receiving the snapshot record, it will be analyzed by DL-based models and/or human experts to extract the features of the scenario, then the "Shadow Mode"~\cite{Silva2022} will be opened on a large number of vehicles to automatically find and record driving scenario data which are highly-similar in feature dimension with the previous corner case. The newly added huge amount data will be feed back to the AD algorithms for model training and validation, facilitating the algorithm's upgrade and improvement. 

Meanwhile, the research community also closely focuses on summarizing the main technology routes and pipelines of autonomous driving.~\cite{10373157,10321736} presented systematic reviews of AD perception algorithms,~\cite{huang2022survey} and~\cite{ma2022verification} presented a comprehensive survey of AD prediction/planning algorithms, respectively. From the perspective of data-driven autonomous driving,~\cite{Hongyang2023,liu2024survey} summarized the development of mainstream autonomous driving datasets. For closed-loop technology,~\cite{zhang2022rethinking} investigated how it could be used for intelligent vehicle planning based on reinforcement learning (RL) methods. However, existing works only focus on one aspect of AD big data technology; none of them conducted a comprehensive review on the whole horizon of data-centric autonomous driving, including the dataset system development, data mining technologies, closed-loop technologies, and empirical studies of data-driven autonomous driving.

Existing challenges strongly call for an in-depth understanding and a systematic review of the whole benchmark data-centric autonomous driving technologies, which prompt us to carry out this survey that integrates both industrial and academic perspectives. The main contributions of this work are summarized in the following:
\begin{itemize}
\item Introducing the first comprehensive taxonomy of autonomous driving datasets categorized by milestone generations, modular tasks, sensor suite, and key features.
\item Providing a systematic review of the state-of-the-art closed-loop data-driven autonomous driving pipelines and related key technologies based on deep learning and generative AI models.
\item Giving empirical studies of how closed-loop big data driven pipeline works in autonomous driving industrial applications.
\item Discussing the pros and cons of current pipelines and solutions, as well as future research direction of data-centric autonomous driving.
\end{itemize}

The systematic organization of this paper is shown in Figure~\ref{fig1}. We start by providing a comprehensive review regarding the development and key features of benchmark autonomous driving datasets in Section~\ref{sec2}. Then, Section~\ref{sec3} introduces the state-of-the-art closed-loop data-driven autonomous driving system, including the pipelines, key technologies, and industrial empirical studies. Following that, the pros and cons of existing methodologies and pipelines are discussed in Section~\ref{sec4}, with guidance on future research directions. Finally, we conclude this paper in Section~\ref{sec5}.

\section{State-of-the-art Autonomous Driving Datasets: Classification and Development} \label{sec2}
The evolution of autonomous driving datasets mirrors the technological advancements and growing ambitions in the field. Early endeavors in the late 20th century, such as the MIT's AVT Research~\cite{MIT-AVT} and UC Berkeley's PATH Program~\cite{PATH}, laid the groundwork with basic sensor data, but were limited by the technology level of the era. There has been a significant leap forward over the last two decades, fueled by advancements in sensor technology, computational power, and sophisticated machine learning algorithms. In 2014, the Society of Automotive Engineers (SAE) announced a systematic six-level (L0-L5) autonomous driving system to the public~\cite{ding2023exploratory}, which has been widely recognized by autonomous driving R\&D progress. Empowered by deep learning, computer vision-based methods have dominated intelligent perception. Deep reinforcement learning and its variants have provided crucial improvements in intelligent planning and decision-making. More recently, Large Language Models (LLMs) and Vision Language Models (VLMs) showcase their strong capability of scene understanding, driving behavior reasoning \& prediction, and intelligent decision making, which open up new possibilities for future development of autonomous driving.

\subsection{Milestone Development of Autonomous Driving Datasets} 
\begin{figure*}[htb]
\centering
\includegraphics[width=0.98\textwidth]{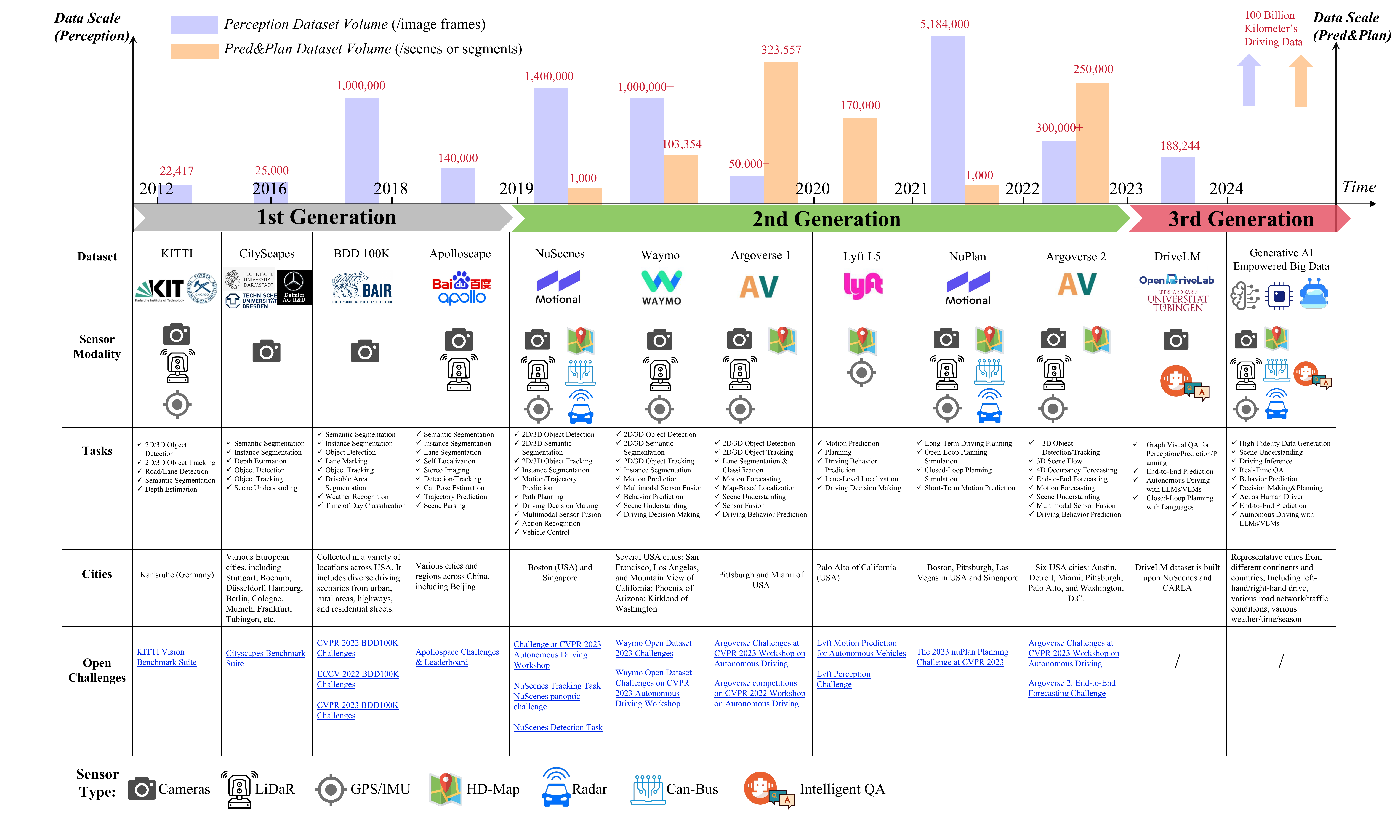}
\caption{The comprehensive illustration of open-source autonomous driving datasets' development characterized by milestone generations. We emphasize the sensor modality, suitable tasks, places of dataset collection, and related challenges.}
\label{fig2}
\end{figure*}

Figure~\ref{fig2} illustrates the milestone development of the open-source autonomous driving datasets following the chronological order. Notable advancements have led to the categorization of mainstream datasets into three identical generations, marked by significant leaps in dataset complexity, volume, scenario variety, and labeling granularity, propelling the field into new frontiers of technological sophistication. Specifically, the horizontal axis denotes the development timeline. The side header of each row includes dataset name, sensor modality, suitable tasks, places of data collection, and related challenges. To further compare datasets across different generations, we visualize the perception and prediction/planning dataset scales using bar charts with different colors. The early phase, denoted as the 1st generation begin at 2012, was spearheaded by KITTI~\cite{KITTI} and Cityscapes~\cite{2016cityscapes}, which provided high-resolution images for perception tasks and were fundamental in benchmarking progress of visual algorithms. Advancing to the 2nd generation, datasets such as NuScenes~\cite{2020nuscenes}, Waymo~\cite{2020scalability}, Argoverse 1~\cite{argoverse1} introduced a multi-sensor approach, integrating vehicle camera, High-Definition Maps (HD-Map), LiDAR, Radar, GPS, IMU, trajectory, surrounding agents' data together, which was crucial for comprehensive driving environment modeling and decision-making processes.
Moving towards recently, NuPlan~\cite{2021nuplan}, Argoverse 2~\cite{argoverse2}, and Lyft L5~\cite{lyft-l5} have significantly raised the impact bar, offering unprecedented data scales and fostering an ecosystem conducive to cutting-edge research. These datasets, characterized by their vast scale and multi-modal sensor integration, have been instrumental in developing algorithms for perception, prediction, and planning tasks, paving the way for the state-of-the-art End2End or hybrid autonomous driving models. Coming to year 2024, we are now welcoming the 3rd generation of autonomous driving datasets. Empowered by VLMs~\cite{gpt4v-wen,Cui2024WACV}, LLMs~\cite{nuscenesQA,wu2023language}, and other Generative AI technologies, the 3rd generation datasets underscore the industry's commitment to addressing the increasingly complex challenges of autonomous driving, such as data Long-Tail Distribution problem, Out-of-Distribution (OOD) detection, Corner Case Analysis, etc. 

\subsection{Dataset Acquisition, Settings, and Key Features}
Table~\ref{tab1} summarizes the data acquisition and annotation settings of highly influential perception datasets, including driving scenario, sensor suite, and annotation method. Specifically, we report the total number of weather/time-of-day/driving condition categories under dataset scenarios, where weather usually includes sunny/cloudy/foggy/rainy/snow/others (extremely conditions); time-of-day usually includes of morning, afternoon, and evening; driving condition usually consists of urban street, main road, side street, rural area, highway, tunnel, parking, etc. The more diverse the scenario is, the more robust and stronger the dataset will be. We also report the regions of dataset collection, denoted as AS (Asia), EU (Europe), NA(North America), SA (South America), AU (Australia), AF (Africa). It is noticeable that Mapillary is collected across AS/EU/NA/SA/AU/AF, and DAWN is collected from Google \& Bing Image Search Engine. For sensor suite, we investigate camera, LiDAR, GPS\&IMU, and others. FV and SV in Table~\ref{tab1} are the abbreviation of Front-View camera and Street-View camera, respectively. 360$^\circ$ denotes panoramic camera settings, which usually consists of multiple front-view cameras, rare-view cameras, and side-view cameras. We can observe that with the development of AD technology, the sensor types and quantities included in the dataset are increasing, and the data modalities are becoming more diverse. As for dataset labeling, early datasets usually adopt manual labeling methods, while the recent NuPlan, Argoverse 2, and DriveLM adopted auto-labeling technologies for AD big data. We believe the transformation from traditional manual annotation to auto-labeling is a major trend in future data-centric autonomous driving.

For prediction and planning tasks, we summarize the input/output components, sensor suite, scene length, and prediction length of the mainstream datasets in Table~\ref{tab2}. As for motion prediction/forecasting task, the input components usually consist of ego-vehicle historical trajectory, surrounding agent historical trajectory, HD-Map, and traffic state information (i.e., traffic signal status, road ID, stop signs, etc). The target output is several most possible trajectories (such as the Top-5 or Top-10 trajectories) of ego-vehicle and/or surrounding agents in a short period of time. Motion forecasting task often adopts sliding time-window settings, where the whole scene is divided into several shorter time-windows. For instance, NuScenes adopts the past 2-second ground-truth data and HD-Map to predict the next 6-second, whereas Argoverse 2 takes the historical 5-second ground-truth and HD-Map to forecast the future 6-second trajectory. 
NuPlan, CARLA, and ApolloScape are among the most popular datasets for planning task. The input components consist of ego/surrounding vehicle historical trajectories, ego-vehicle motion states, and driving scenario representations. While NuPlan and ApolloScape are acquired in real-world, CARLA is a simulated dataset. CARLA contains road images captured during simulated driving sessions in different towns. Each road image i accompanied with a steering angle, which represents the required adjustment to keep the vehicle on normal driving. The prediction length of planning can be varied according to the requirements of different algorithms.
\begin{table*}[htb]
\centering
\tiny
\begin{threeparttable}
\resizebox{0.95\linewidth}{!}{
\begin{tabular}{c|c|cccc|ccccc|c|c}
\hline
\multirow{2}{*}{\textbf{Dataset}} & \multirow{2}{*}{\textbf{Year}} & \multicolumn{4}{c|}{\textbf{Dataset Scenario}}                                & \multicolumn{5}{c|}{\textbf{Sensor Suite}}                                                                                                                                                                              & \multirow{2}{*}{\textbf{Annotation}} & \multirow{2}{*}{\textbf{Auto-labeling}} \\ \cline{3-11}
                                  &                                & \multicolumn{1}{c|}{Weather} & \multicolumn{1}{c|}{Time-of-day} & \multicolumn{1}{c|}{\begin{tabular}[c]{@{}c@{}}Drive \\ Condition\end{tabular}} & Region  & \multicolumn{1}{c|}{Camera}      & \multicolumn{1}{c|}{LiDAR}                       & \multicolumn{1}{c|}{GPS\&IMU}                    & \multicolumn{1}{c|}{CAN-bus}                     & HD-Map                      &                                      &                                         \\ \hline
KITTI\tnote{[1]}                   & 2012                           & \multicolumn{1}{c|}{1}       & \multicolumn{1}{c|}{1}           & \multicolumn{1}{c|}{3}                                                          & EU      & \multicolumn{1}{c|}{FV}          & \multicolumn{1}{c|}{\Checkmark}   & \multicolumn{1}{c|}{\Checkmark}   & \multicolumn{1}{c|}{\XSolidBrush} & \XSolidBrush & 2D b-box \&3D b-box                  & \XSolidBrush             \\ \hline
CityScapes\tnote{[2]}               & 2016                           & \multicolumn{1}{c|}{1}       & \multicolumn{1}{c|}{1}           & \multicolumn{1}{c|}{3}                                                          & EU      & \multicolumn{1}{c|}{FV}          & \multicolumn{1}{c|}{\XSolidBrush} & \multicolumn{1}{c|}{\XSolidBrush} & \multicolumn{1}{c|}{\XSolidBrush} & \XSolidBrush & 2D Seg                               & \XSolidBrush             \\ \hline
Mapillary\tnote{[3]}            & 2016                           & \multicolumn{1}{c|}{5}       & \multicolumn{1}{c|}{3}           & \multicolumn{1}{c|}{3}                                                          & Global  & \multicolumn{1}{c|}{SV}          & \multicolumn{1}{c|}{\XSolidBrush} & \multicolumn{1}{c|}{\XSolidBrush} & \multicolumn{1}{c|}{\XSolidBrush} & \XSolidBrush & 2D Seg                               & \XSolidBrush             \\ \hline
Apolloscape\tnote{[4]}           & 2016                           & \multicolumn{1}{c|}{4}       & \multicolumn{1}{c|}{3}           & \multicolumn{1}{c|}{3}                                                          & AS      & \multicolumn{1}{c|}{FV}          & \multicolumn{1}{c|}{\XSolidBrush} & \multicolumn{1}{c|}{\Checkmark}   & \multicolumn{1}{c|}{\XSolidBrush} & \XSolidBrush & 3D b-box \& 2D Seg                   & \XSolidBrush             \\ \hline
BDD\tnote{[5]}                    & 2018                           & \multicolumn{1}{c|}{5}       & \multicolumn{1}{c|}{2}           & \multicolumn{1}{c|}{4}                                                          & NA      & \multicolumn{1}{c|}{FV}          & \multicolumn{1}{c|}{\XSolidBrush} & \multicolumn{1}{c|}{\XSolidBrush} & \multicolumn{1}{c|}{\XSolidBrush} & \XSolidBrush & Text                               & \XSolidBrush             \\ \hline
SemanticKITTI\tnote{[6]}         & 2019                           & \multicolumn{1}{c|}{1}       & \multicolumn{1}{c|}{1}           & \multicolumn{1}{c|}{4}                                                          & EU      & \multicolumn{1}{c|}{\_\_}           & \multicolumn{1}{c|}{\Checkmark}   & \multicolumn{1}{c|}{\XSolidBrush} & \multicolumn{1}{c|}{\XSolidBrush} & \XSolidBrush & 3D Seg                               & \XSolidBrush             \\ \hline
DAWN\tnote{[7]}                   & 2019                           & \multicolumn{1}{c|}{6}       & \multicolumn{1}{c|}{3}           & \multicolumn{1}{c|}{3}                                                          & Unknown & \multicolumn{1}{c|}{FV}          & \multicolumn{1}{c|}{\XSolidBrush} & \multicolumn{1}{c|}{\XSolidBrush} & \multicolumn{1}{c|}{\XSolidBrush} & \XSolidBrush & 2D b-box                             & \XSolidBrush             \\ \hline
UNDD\tnote{[8]}                    & 2019                           & \multicolumn{1}{c|}{2}       & \multicolumn{1}{c|}{2}           & \multicolumn{1}{c|}{2}                                                          & Unknown & \multicolumn{1}{c|}{FV}          & \multicolumn{1}{c|}{\XSolidBrush} & \multicolumn{1}{c|}{\XSolidBrush} & \multicolumn{1}{c|}{\XSolidBrush} & \XSolidBrush & 2D Seg                          & \XSolidBrush             \\ \hline
NuScenes\tnote{[9]}              & 2019                           & \multicolumn{1}{c|}{3}       & \multicolumn{1}{c|}{2}           & \multicolumn{1}{c|}{4}                                                          & AS,NA   & \multicolumn{1}{c|}{360$^\circ$} & \multicolumn{1}{c|}{\Checkmark}   & \multicolumn{1}{c|}{\Checkmark}   & \multicolumn{1}{c|}{\Checkmark}   & \Checkmark   & 3D b-box \& 3D Seg                   & \XSolidBrush             \\ \hline
Argoverse 1\tnote{[10]}           & 2019                           & \multicolumn{1}{c|}{2}       & \multicolumn{1}{c|}{1}           & \multicolumn{1}{c|}{1}                                                          & NA      & \multicolumn{1}{c|}{360$^\circ$} & \multicolumn{1}{c|}{\Checkmark}   & \multicolumn{1}{c|}{\XSolidBrush} & \multicolumn{1}{c|}{\XSolidBrush} & \Checkmark   & 3D b-box \& 3D Seg                   & \XSolidBrush             \\ \hline
Waymo\tnote{[11]}                & 2019                           & \multicolumn{1}{c|}{2}       & \multicolumn{1}{c|}{3}           & \multicolumn{1}{c|}{5}                                                          & NA      & \multicolumn{1}{c|}{360$^\circ$} & \multicolumn{1}{c|}{\Checkmark}   & \multicolumn{1}{c|}{\Checkmark}   & \multicolumn{1}{c|}{\XSolidBrush} & \XSolidBrush & 2D b-box \&3D b-box                  & \XSolidBrush             \\ \hline
KITTI-360\tnote{[12]}              & 2020                           & \multicolumn{1}{c|}{1}       & \multicolumn{1}{c|}{1}           & \multicolumn{1}{c|}{1}                                                          & EU      & \multicolumn{1}{c|}{360$^\circ$} & \multicolumn{1}{c|}{\Checkmark}   & \multicolumn{1}{c|}{\XSolidBrush} & \multicolumn{1}{c|}{\XSolidBrush} & \XSolidBrush & 3D b-box \& 3D Seg                   & \XSolidBrush             \\ \hline
ONCE\tnote{[13]}                 & 2021                           & \multicolumn{1}{c|}{3}       & \multicolumn{1}{c|}{2}           & \multicolumn{1}{c|}{5}                                                          & AS      & \multicolumn{1}{c|}{360$^\circ$} & \multicolumn{1}{c|}{\Checkmark}   & \multicolumn{1}{c|}{\XSolidBrush} & \multicolumn{1}{c|}{\XSolidBrush} & \XSolidBrush & 3D b-box                             & \XSolidBrush             \\ \hline
NuPlan\tnote{[14]}                 & 2021                           & \multicolumn{1}{c|}{3}       & \multicolumn{1}{c|}{2}           & \multicolumn{1}{c|}{4}                                                          & AS,NA   & \multicolumn{1}{c|}{360$^\circ$} & \multicolumn{1}{c|}{\Checkmark}   & \multicolumn{1}{c|}{\Checkmark}   & \multicolumn{1}{c|}{\Checkmark}   & \Checkmark   & 3D b-box                             & \Checkmark               \\ \hline
Argoverse 2\tnote{[15]}           & 2022                           & \multicolumn{1}{c|}{2}       & \multicolumn{1}{c|}{1}           & \multicolumn{1}{c|}{1}                                                          & NA      & \multicolumn{1}{c|}{360$^\circ$} & \multicolumn{1}{c|}{\Checkmark}   & \multicolumn{1}{c|}{\XSolidBrush} & \multicolumn{1}{c|}{\XSolidBrush} & \Checkmark   & 3D b-box                             & \Checkmark               \\ \hline
DriveLM\tnote{[16]}               & 2023                           & \multicolumn{1}{c|}{3}       & \multicolumn{1}{c|}{2}           & \multicolumn{1}{c|}{4}                                                          & AS,NA   & \multicolumn{1}{c|}{360$^\circ$} & \multicolumn{1}{c|}{\XSolidBrush} & \multicolumn{1}{c|}{\XSolidBrush} & \multicolumn{1}{c|}{\XSolidBrush} & \XSolidBrush & graph visual QA                      & \Checkmark               \\ \hline
\end{tabular}
}
\begin{tablenotes}
\item {[1]}~\cite{KITTI}, {[2]}~\cite{2016cityscapes}, {[3]}~\cite{Mapillary}, {[4]}~\cite{apolloscape}, {[5]}~\cite{bdd100k}, {[6]}~\cite{semantickitti}, {[7]}~\cite{kenk2020dawn}, {[8]}~\cite{nag2019s}, {[9]}~\cite{2020nuscenes}, {[10]}~\cite{argoverse1}, {[11]}~\cite{2020scalability}, {[12]}~\cite{liao2022kitti}, {[13]}~\cite{mao2021one}, {[14]}~\cite{2021nuplan}, {[15]}~\cite{argoverse2}, {[16]}~\cite{sima2023drivelm}
\end{tablenotes}
\end{threeparttable}
\caption{Summary of the data acquisition and annotation settings of highly influential perception datasets}
\label{tab1}
\end{table*}

\begin{table*}[htb]
\centering
\tiny
\begin{threeparttable}
\resizebox{0.85\linewidth}{!}{
\begin{tabular}{c|l|l|llll|c|c}
\hline
\multirow{2}{*}{\textbf{Dataset}}                                                       & \multicolumn{1}{c|}{\multirow{2}{*}{\textbf{Input components}}}                                                                                          & \multicolumn{1}{c|}{\multirow{2}{*}{\textbf{Target}}}                                                       & \multicolumn{4}{c|}{\textbf{Sensor Suite}}                                                                                                                                                                 & \multirow{2}{*}{\textbf{Scene Length}} & \multirow{2}{*}{\textbf{\begin{tabular}[c]{@{}c@{}}Pred Trajectory \\ Length\end{tabular}}} \\ \cline{4-7}
                                                                                        & \multicolumn{1}{c|}{}                                                                                                                                    & \multicolumn{1}{c|}{}                                                                                       & \multicolumn{1}{c|}{GPS \& IMU}  & \multicolumn{1}{c|}{HD-Map}      & \multicolumn{1}{c|}{Camera}      & \multicolumn{1}{c|}{Others}                                                                       &                                        &                                                                                             \\ \hline
\begin{tabular}[c]{@{}c@{}}Argoverse 2 \\ Motion\\  Forecasting \\ Dataset\tnote{[1]}\end{tabular} & \begin{tabular}[c]{@{}l@{}}Ego-vehivle traj, \\ Surrounding agent traj, \\ HD-Map, \\ Traffic state info\end{tabular}                                    & \begin{tabular}[c]{@{}l@{}}Multiple possible \\ trajectories of\\ ego/surrounding \\ vehicles\end{tabular}  & \multicolumn{1}{c|}{\Checkmark}   & \multicolumn{1}{c|}{\Checkmark}   & \multicolumn{1}{c|}{\XSolidBrush} & \begin{tabular}[c]{@{}l@{}}BEV centroid and \\ heading angle \\ of every track\end{tabular}       & 11s                                    & 6s                                                                                          \\ \hline
\begin{tabular}[c]{@{}c@{}}NuScenes \\ Prediction\\  Dataset\tnote{[2]}\end{tabular}               & \begin{tabular}[c]{@{}l@{}}Ego-vehivle traj, \\ Surrounding agent traj, \\ HD-Map, Traffic state info\end{tabular}                                       & \begin{tabular}[c]{@{}l@{}}Multiple possible \\ trajectories of \\ ego/surrounding \\ vehicles\end{tabular} & \multicolumn{1}{c|}{\Checkmark}   & \multicolumn{1}{c|}{\Checkmark}   & \multicolumn{1}{c|}{\XSolidBrush} & Can-bus info                                                                                      & 20s                                    & 6s                                                                                          \\ \hline
\begin{tabular}[c]{@{}c@{}}Waymo Motion \\ Prediction Dataset\tnote{[3]}\end{tabular}              & \begin{tabular}[c]{@{}l@{}}Ego-vehivle traj, \\ Surrounding agent traj, \\ HD-Map, Traffic state info\end{tabular}                                       & \begin{tabular}[c]{@{}l@{}}Multiple possible \\ trajectories of \\ ego/surrounding \\ vehicles\end{tabular} & \multicolumn{1}{c|}{\Checkmark}   & \multicolumn{1}{c|}{\Checkmark}   & \multicolumn{1}{c|}{\XSolidBrush} & \begin{tabular}[c]{@{}l@{}}Agent motion states\\  (velocity, heading, \\ valid flag)\end{tabular} & 20s                                    & 8s                                                                                          \\ \hline
\begin{tabular}[c]{@{}c@{}}ApolloScape \\ Trajectory \\ Dataset\tnote{[4]}\end{tabular}            & \begin{tabular}[c]{@{}l@{}}Ego-vehicle traj, \\ Ego-vehicle motion states, \\ Driving scenario info\end{tabular}                                         & \begin{tabular}[c]{@{}l@{}}Trajectory planning \\ for ego vehicle\end{tabular}                              & \multicolumn{1}{c|}{\Checkmark}   & \multicolumn{1}{c|}{\XSolidBrush} & \multicolumn{1}{c|}{\XSolidBrush} & Heading angle                                                                                     & 60s                                    & 3s                                                                                          \\ \hline
NuPlan\tnote{[5]}                                                                    & \begin{tabular}[c]{@{}l@{}}Ego-vehicle traj, \\ Ego-vehicle motion states, \\ Surrounding vehicle traj, \\ Driving scenario info, \\ HD-Map\end{tabular} & \begin{tabular}[c]{@{}l@{}}Trajectory planning \\ for ego vehicle\end{tabular}                              & \multicolumn{1}{c|}{\Checkmark}   & \multicolumn{1}{c|}{\Checkmark}   & \multicolumn{1}{c|}{\Checkmark}   & \begin{tabular}[c]{@{}l@{}}Steering inputs, \\ CAN-bus info\end{tabular}                          & \_\_                                  & 2s                                                                                          \\ \hline
CARLA\tnote{[6]}                                                                    & \begin{tabular}[c]{@{}l@{}}Ego-vehicle traj, \\ Ego-vehicle motion states, \\ Driving scenario info\end{tabular}                                         & \begin{tabular}[c]{@{}l@{}}Trajectory planning \\ for ego vehicle\end{tabular}                              & \multicolumn{1}{c|}{\XSolidBrush} & \multicolumn{1}{c|}{\XSolidBrush} & \multicolumn{1}{c|}{\Checkmark}   & Steering inputs                                                                                   & \_\_                                    & \_\_                                                                                       \\ \hline
\end{tabular}
}
\begin{tablenotes}
\item {[1]}~\cite{argoverse2}, {[2]}~\cite{2020nuscenes}, {[3]}~\cite{ettinger2021}, {[4]}~\cite{apolloscape}, {[5]}~\cite{2021nuplan}, {[6]}~\cite{2017carla}
\end{tablenotes}
\end{threeparttable}
\caption{Summary of the input components, target, sensor suite, scene \& prediction length of popular prediction/planning datasets}
\label{tab2}
\end{table*}

Following that, we select one of the most representative datasets from each generation (KITTI~\cite{KITTI}, NuScenes~\cite{2020nuscenes}, DriveLM~\cite{sima2023drivelm}) and provide an introduction to the related dataset details. \\
\noindent \textbf{KITTI Dataset.} KITTI represents a paradigm shift in utilizing visual sensors for tasks where traditional reliance was on GPS, laser rangefinders, and radar. KITTI provides a rich array of calibrated and synchronized data, as shown in Table~\ref{tab1}. Collected from diverse urban (Karlsruhe city), rural, and highway, this dataset focuses on developing benchmarks for stereo, optical flow, visual odometry/SLAM, and 3D object detection, diverging from conventional datasets that often lack realism or comprehensive 3D information.

\noindent \textbf{NuScenes Dataset.} Drawing inspiration from KITTI, NuScenes stands out as the first collection providing comprehensive data from AV's entire sensor suite, including 6xcameras, 1xLIDAR, 5xRADARs, along with GPS \& IMU. It surpasses KITTI with 7 times increase in object annotations. Unlike previous datasets that solely concentrated on perception, NuScenes offers a holistic view of the entire AD tasks, including perception, motion forecasting, decision making and planning, making the End2End DL-based pipeline possible. NuScenes encompasses 1000 meticulously chosen driving scenarios from Boston and Singapore, rich in diversity. Each 20-second scene captures a range of driving maneuvers and unique traffic challenges.

\noindent \textbf{DriveLM Dataset.} 
Autonomous driving systems are progressing, yet they still lack critical capabilities essential for robust operation in the real world, such as accurately handling unseen scenarios or unfamiliar objects, interacting with human drivers, providing explanations in decision-making progress. As a pioneer of exploring the 3rd generation's AD big data, DriveLM aims to tackle these challenges by incorporating VLMs, LLMs, and Graph Visual Question Answering (GVQA) technologies into the AD stack. DriveLM leverages these strengths to foster generalization in planning and to facilitate human-robot interaction through models that can understand and generate natural language descriptions. This compatibility with human communication is a key differentiator from current methods that rely on purely end-to-end differentiable architectures. DriveLM is particularly innovative in the design of thinking chain GVQA under the AD context. GVQA involves formulating Pi-3 reasoning—consisting of a series of question-answer pairs—to simulate the human reasoning process involved in driving. Aside from that, DirveLM also introduces dedicated metrics for assessing performance in a more logical and reasoned manner.

\section{Closed-Loop Data-Driven Autonomous Driving System} \label{sec3}
We're now shifting from the previous era of software \& algorithm defined autonomous driving towards the new inspiring era of big data-driven \& intelligent model collaborative autonomous driving. Closed-loop data-driven systems aim to bridge the gap between AD algorithm training and their real-world application/deployment. Unlike traditional open-loop methods, where models are passively trained on datasets collected from human client driving or road testing, closed-loop systems interact dynamically with the real environment. This approach addresses the distribution shifting challenge--where behavior learned from static datasets may not translate to the dynamic nature of real-world driving scenarios. Closed-loop systems allow AVs to learn from interactions and adapt to new situations, improving through iterative cycles of action and feedback. 

However, building real-world closed-loop data-centric AD systems remain challenging because of several critical issues: The first issue is associated with AD data collection. In real-world data acquisition, the majority of data samples are common/normal driving scenarios, whereas corner case and abnormal driving scenario data can hardly ever be acquired. Second, further endeavors are needed to explore accurate and efficient AD data auto-labeling methods. Third, in order to alleviate the problem of AD model's underperformance in certain scenarios within urban environments, scene data mining and scene understanding should be emphasized. 

\subsection{State-of-the-art Closed-Loop Autonomous Driving Pipelines}
The autonomous driving industry is actively constructing integrated big data platforms in response to the challenges caused by the accumulation of massive amounts of AD data. This can be aptly termed as the new infrastructure in the era of data-driven autonomous driving. During our investigation of the data-driven closed-loop systems developed by top-tier AD companies/research institutions, we have identified several commonalities:
\begin{itemize}
\item These pipelines usually follow a workflow circle that includes: (I) data acquisition, (II) data storage, (III) data selection \& preprocessing, (IV) data labeling, (V) AD model training, (VI) simulation/test validation, and (VII) real-world deployment. 
\item For the design of closed-loops within the system, existing solutions either choose separately set "Data Close-Loop" \& "Model Close-Loop", or separately set cycles for different stages: "Close Loop during R\&D stage" and "Close Loop during deployment stage". 
\item Aside from that, the industry also emphasizes the long-tail distribution problem of real-world AD datasets and the challenges when dealing with corner case. Tesla and NVIDIA are industry pioneers in this realm, and their data system architectures offer significant reference for the development of the field.
\end{itemize}

\begin{figure*}[htb]
\centering
\includegraphics[width=0.98\textwidth]{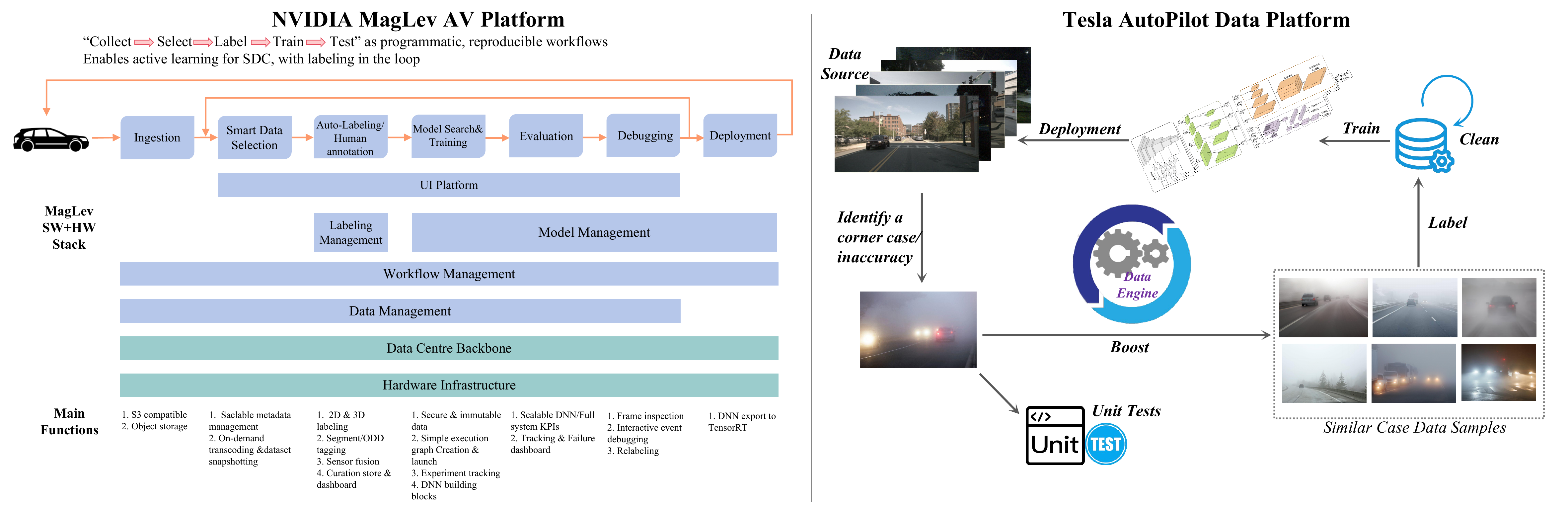}
\caption{The workflow illustration of two pioneer data-driven closed-loop autonomous driving pipelines: NVIDIA's MagLev AV Platform (left) and Tesla AutoPilot Data Platform (right).}
\label{fig3}
\end{figure*}

NVIDIA MagLev AV platform~\cite{opML-NV} (Figure~\ref{fig3}(left)) follows the "collect $\to$ select $\to$ label $\to$ train $\to$ test" as the programmatic, it is a reproducible workflow which enables active learning for SDC, with intelligent labeling in the loop. MagLev primarily includes two closed-loop pipelines. The first loop is autonomous driving data-centric, which starts with data ingestion and smart selection, moving through labeling and annotation, followed by model search and training. The trained models are then evaluated, debugged, and finally deployed in real-world. The second closed-loop is the platform's infrastructure support system, which includes data center backbone and hardware infrastructure. This loop includes secure data handling, scalable DNN and system KPIs, dashboard for tracking and debugging. It supports the full cycle of AV development, ensuring continuous improvement and integration of real-world data and simulation feedback into the development process.

Tesla AutoPilot Data Platform~\cite{wav2023T} (Figure~\ref{fig3} (right)) is another representative AD platform that emphasizes using big data-driven closed-loop pipelines to significantly boost autonomous driving model performance. The pipeline starts from source data collection, which usually comes from Tesla's fleet learning, event-trigger vehicle-end data collection, and shadow mode. The collected data will be stored, managed, and inspected by data platform algorithms or human experts. Whenever a corner case/inaccuracy is identified, the data engine will retrieve and match data samples that are highly similar to the corner case/inaccuracy event from the existing database. At the same time, units testing will be developed to replicate the scenario and rigorously test the system's responses. Following that, the retrieved data samples will be labeled by either auto-labeling algorithms or human experts. Then, well-labeled data will be feedback to AD database, the database will be updated to generate new versions of training dataset for AD perception/prediction/planning/control models. After model training, validation, simulation and real-world testing, the new AD models with higher performance will be released and deployed. 

\subsection{High-Fidelity AD Data Generation and Simulation based on Generative AI}
The majority of AD data samples collected from real-world acquisition are common/normal driving scenarios, of which we already have a massive amount of similar samples in the database. However, to collect a certain type of AD data samples from real-world acquisition, we need to drive for exponentially long hours, which is not feasible in industrial applications. As such, methods for high-fidelity autonomous driving data generation and simulation have aroused significant attention from academia. 
CARLA~\cite{2017carla,2021kitti-carla} is an open-source simulator for autonomous driving research that enables the generation of autonomous driving data under various settings specified by the user. The advantage of CARLA lies in its flexibility, allowing users to create diverse road conditions, traffic scenarios, and weather dynamics, which facilitates comprehensive model training and testing. However, as a simulator, its primary disadvantage lies in the domain gap. AD data generated by CARLA cannot fully mimic the real-world physics and visual effects; the dynamic and complex characteristics of real driving environment are also not represented.

More recently, World Model~\cite{ha2018world,lin2020improving}, with its more advanced intrinsic concept and more promising performance, has been employed for high-fidelity AD data generation. \textbf{World Model can be defined as} an AI system that builds an internal representation of the environment it perceives, and uses the learned representation to simulate data or events within the environment. The objective of general world models are to represent and simulate various situations and interactions, like a mature human will encountered in the real-world. In autonomous driving fields, GAIA-1~\cite{hu2023gaia} and DriveDreamer~\cite{drivedreamer} are representative works for data generation based on world models. GAIA-1 is a generative AI model that enables image/video to image/video generation by taking the original image/video along with text and action prompts as input. The input modalities of GAIA-1 are encoded into a unified sequence of tokens. These tokens are processed by an autoregressive transformer within a world model to predict subsequent image tokens. A video decoder then reconstructs these tokens into a coherent video output with enhanced temporal resolution, enabling dynamic and contextually rich visual content generation. DriveDreamer innovatively employs a diffusion model within its architecture, focusing on capturing the intricacies of real-world driving environments. Its two-stage training pipeline first enables the model to learn structured traffic constraints and then progresses to forecasting future states, ensuring a robust environmental understanding tailored for autonomous driving applications.

\subsection{Auto-Labeling Methods for Autonomous Driving Datasets}
\begin{figure*}[tb]
\centering
\includegraphics[width=0.99\textwidth]{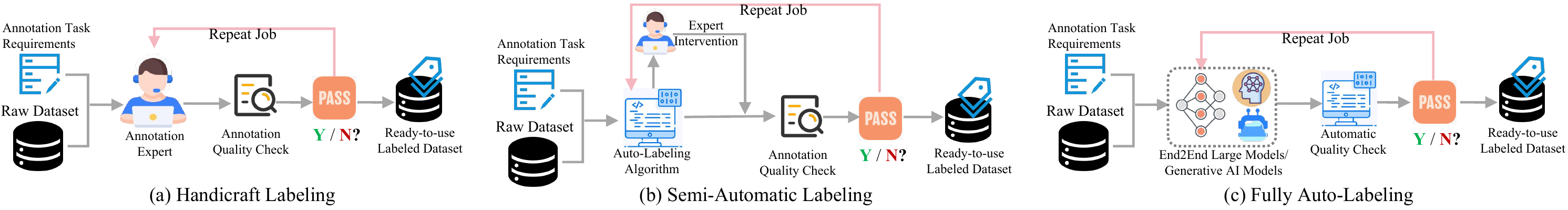}
\caption{The detailed workflows of mainstream AD data labeling pipelines. AD data labeling is usually task/model specific, with pre-defined requirements. It's usually not a one-time task, but a cyclical procedure.}
\label{fig5}
\end{figure*}

High-quality data annotation is indispensable for the success and reliability of AD algorithms. Up to date, the data labeling pipelines can be characterized as three types, ranging from traditional handicraft labeling to semi-automatic labeling and then to the most advanced fully auto-labeling approach, as depicted in Figure~\ref{fig5}. AD data labeling is usually considered as task/model specific. The workflow starts with carefully preparing the requirements of annotation task and raw dataset. Then, the next step is to generate initial annotation results using human experts, auto-labeling algorithms, or End2End large models. Following that, the annotation quality will be checked by either human experts or automatic quality-check algorithms, according to the pre-defined requirements. If the annotation results in this round fail to pass quality check, they will be sent back to the labeling loop again and repeat this annotation job, until they meet the pre-defined requirements. Finally, we can obtain the ready-to-use labeled AD dataset.

Auto-labeling methods are pivotal for closed-loop autonomous driving big data platforms in alleviating the intensive labor of manual annotation, improving the efficiency of AD data closed-loop cycle, and reducing the related costs. Classic auto-labeling tasks include scene classification \& understanding~\cite{9913352}. Recently, with the popularization of Bird's-Eye-View (BEV) perception methodology~\cite{10321736}, the industrial standard for AD data labeling is also increasing, and the auto-labeling tasks are becoming more complicated. In the scenario of today's industrial frontier, 3D dynamic object auto-labeling and 3D static scene auto-labeling are two commonly mentioned advanced auto-labeling tasks.

Scene classification and understanding are foundational in autonomous driving big data platform, where the system categorizes video frames into predefined scenarios, such as driving places (street, highway, urban overpass, main road, etc) and scene weathers (sunny, rain, snow, foggy, thunderstorm, etc). CNN-based methods are usually employed for scene classification, which include Pre-training+Fine-tune CNN models~\cite{liu2019novel}, multi-view and multi-layer CNN models~\cite{terven2023}, and various CNN-based models for improving scene representation~\cite{dixit2016object,chen2020scene}. Scene understanding~\cite{2023openscene,YOLOv8} goes beyond mere classification. It involves interpreting dynamic elements within the scene, such as surrounding vehicle agents, pedestrian, and traffic lights. Aside from image-based scene understanding, LiDAR-based data sources such as SemanticKITTI~\cite{semantickitti} have also been widely adopted because of the fine-grained geometric information they provide. 

The emergence of 3D dynamic object auto-labeling and 3D static scene auto-labeling are to meet the the requirements of the widely adopted BEV perception technology. Waymo~\cite{qi2021offboard} proposed a 3D Auto-Labeling pipeline from LiDAR point cloud sequential data, which employs a 3D detector to localize objects frame by frame. The identified objects' bounding boxes across the frames are then linked through a multi-object tracker. Object track data (the corresponding point clouds + 3D bounding boxes at every frame) are extracted for each object and undergone the object-centric auto-labeling with a divide-and-conquer architecture to generate the final refined 3D bounding boxes as labels. The Auto4D pipeline~\cite{yang2021auto4d} proposed by Uber first explores AD perception labeling under spatial-temporal scale. \textit{3D object bounding box labeling} within the spatial scale, along with \textit{1D corresponded timestamp labeling} within the temporal scale, is termed as \textbf{4D labeling} in autonomous driving field. Auto4D pipeline begins with sequential LiDAR point clouds to establish an initial object trajectory. This trajectory is refined by the Object Size Branch, which uses object observations to encode and decode object size. Concurrently, the Motion Path Branch encodes path observations and motion, allowing the Path Decoder to refine the trajectory with constant object size.

3D static scene auto-labeling can be considered as HD-Map generation, where lanes, road boundaries, crosswalks, traffic lights, and other related elements in the driving scene should be annotated. There are several attractive research works under this topic: vision-based methods such as MV-Map~\cite{xie2023mv}, NeMO~\cite{zhu2023nemo}; LiDAR-based methods such as VMA~\cite{chen2023vma}; Pre-training 3D scene reconstruction methods such as Occ-BEV~\cite{min2023occ}, OccNet~\cite{tong2023scene}, AD-PT~\cite{yuan2023ad}, ALSO~\cite{boulch2023also}. VMA is a recently proposed work for 3D static scene auto-labeling. VMA framework utilizes crowd-sourced, multi-trip aggregated LiDAR point clouds to reconstruct static scenes, segmenting them into units for processing. The MapTR-based unit annotator encodes the raw inputs into feature maps, through querying and decoding, yielding semantically typed point sequences. The outputs of VMA are vectorized maps, they will be refined via closed-loop annotation and human verification, resulting in satisfied HD-Maps for autonomous driving. 


\subsection{Empirical Study}

\begin{figure}[thb]
\centering
\includegraphics[width=0.99\columnwidth]{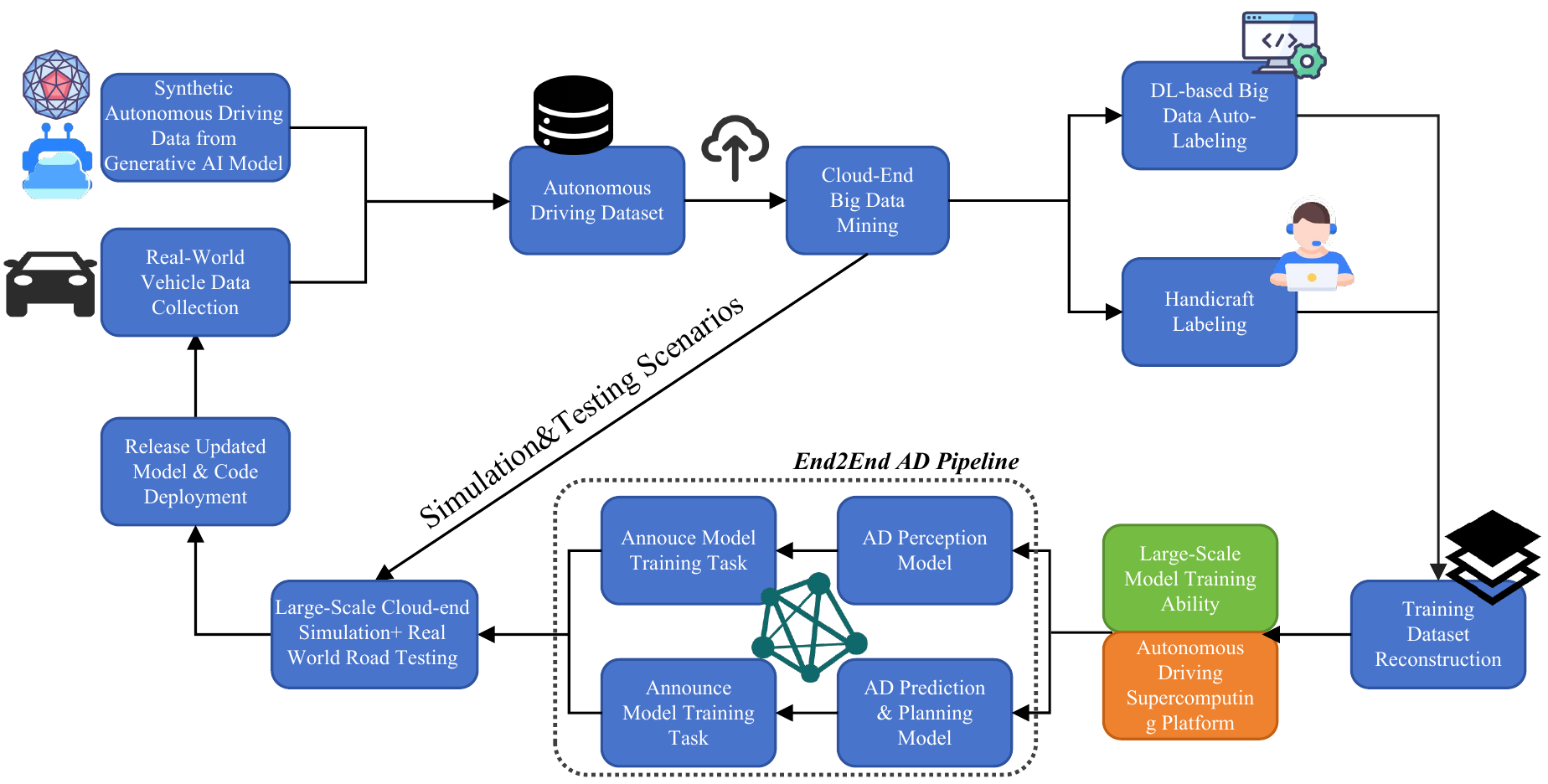}
\caption{The real-world application diagram of advanced closed-loop autonomous driving big data platform.}
\label{fig4}
\end{figure}
We provide an empirical study to better illustrate the advanced closed-loop AD data platform mentioned in this work. The overall procedure diagram is depicted in Figure~\ref{fig4}. The goal of researchers in this case is to develop an AD big data closed-loop pipeline based on Generative AI and various deep learning-based algorithms, so as to achieve data close loop in both autonomous driving algorithm R\&D phase and OTA upgrade phase (after real-world deployment). Specifically, generative AI models are used for (1) generating scene-specific high-fidelity AD data based on the text prompts provided by engineers. (2) AD big data auto-labeling for efficiently preparing the ground-truth labels. 

There are two closed-loops shown in the diagram. The large one is for autonomous driving algorithm R\&D phase, which starts with data collection of both synthetic autonomous driving data from generative AI models and data samples acquired from real-world driving. The two kinds of data sources are integrated as an autonomous driving dataset that is mined at the cloud-end for valuable insights. Following that, the dataset enters a dual path of labeling: either deep learning-based auto-labeling or manual handcraft labeling, ensuring speed and precision in annotations. The labeled data is then used to train models on a high-capacity autonomous driving supercomputing platform. These models are subjected to simulation and real-world road testing to evaluate their efficacy, leading to the release of an autonomous driving model and subsequent deployment. The small one is for the OTA upgrade phase after real-world deployment, which involves large-scale cloud-end simulations and real-world testing to collect inaccuracy/corner case of the AD algorithm. The identified inaccuracy/corner case is used to inform the next iteration of model testing and updating. For example, suppose we find our AD algorithms perform poorly in tunnel driving scenario. The identified tunnel driving corner case will be announced to the loop immediately, and updated in the next iteration. The generative AI model will take descriptions related to tunnel driving scenario as text prompts to generate large-scale of tunnel driving data samples. The generated data together with original dataset will be fed into simulation, testing, and model updating. The iterative nature of these processes is essential for refining the models to adapt to challenging environments and new data, maintaining a high level of accuracy and reliability in autonomous driving functionalities.

\section{Discussions} \label{sec4}
\noindent \textbf{New Autonomous Driving Datasets at 3rd Generation and Beyond.} Despite the success of Foundation Models such as LLMs/VLMs in language understanding and computer vision, directly applying them to autonomous driving remains to be challenging. The reasons are two-folds: For one thing, these LLMs/VLMs must have the capability to comprehensively integrate and understand multi-source AD big data (e.g., FOV images/videos, LiDAR cloud-points, HD-Map, GPS/IMU data, etc), which is more difficult than comprehending the images we seen in daily life. For another, existing data scale and quality in autonomous driving field is not comparable to other fields (e.g., financial and medical), making it difficult to support the training and optimization of larger volume LLMs/VLMs. The current scale and quality of autonomous driving big data is limited because of regulations, privacy concerns, and cost. We believe that with the joint efforts of all parties, the next generation's AD big data will have significant improvements in both scale and quality. 

\noindent \textbf{Hardware Supports for Autonomous Driving Algorithms.} Current hardware platforms have made significant strides, particularly with the advent of specialized processors like GPUs and TPUs that offer substantial parallel computation capabilities essential for deep learning tasks. High-performance computing resources, both onboard vehicles and in cloud infrastructures, are critical for processing the vast data streams generated by vehicular sensors in real-time. Despite these advancements, there remain limitations in scalability, energy efficiency, and processing speeds when dealing with the increasing complexity of autonomous driving algorithms. VLMs/LLMs guided user-vehicle interaction is a very promising application case. It could be possible to collect user-specific behavior big data based on this application. However, the equipment of VLMs/LLMs on vehicle-end will request high-standard hardware computing resources, and the interactive application is expected to be low latency. As such, there could be some light-weight large models for autonomous driving in the future, or the compression technology of LLMs/VLMs will be further investigated.

\noindent \textbf{Personalized Autonomous Driving Recommendation based on User Behavior Data.} Intelligent vehicle, has evolved from simple means of transportation to the latest application extensions in intelligent terminal scenarios. Therefore, people's expectation for vehicles equipped with advanced autonomous driving functionalities is that they are capable of learning the driver's behavioral preferences, such as driving styles and traveling route preferences, from their historical driving data records. This would enable intelligent vehicles to better align with user favorites when assisting drivers in vehicle control, driving decision-making, and route planning in the future. We term the conception described above as the \textit{Personalized Autonomous Driving Recommendation Algorithm}. Recommendation systems have been widely employed in e-commerce online shopping, food delivery, social media, and live-streaming platforms. However, in the field of autonomous driving, personalized recommendation is still in its infancy stage. We believe in the near future, a more suitable data system and data acquisition mechanism will be designed to collect big data on user driving behavior preferences, with users' permission and compliance with related regulations, so as to achieve customized autonomous driving recommendation system for users. 

\noindent \textbf{Data Security and Trustworthy Autonomous Driving.} The vast collection of autonomous driving big data poses significant challenges for data security and user privacy protection. As vehicles become increasingly connected with the advancing of Connected-Autonomous-Vehicles (CAV) and Internet-of-Vehicles (IoV) technologies, the collection of detailed user data, from driving habits to frequent routes, raises concerns about the potential misuse of personal information. We suggest the necessity of transparency in the collected data types, retention policies, and third-party sharing. It stresses the importance of user consent and control, including respecting "do not track" requests and providing options to delete personal data. For the autonomous driving industry, safeguarding this data while fostering innovation requires stringent adherence to these guidelines, ensuring user trust and compliance with evolving privacy legislation.

Aside from data security and privacy, another concern is how to realize trustworthy autonomous driving. With the vast development of AD technology, intelligent algorithms and generative AI models (e.g., LLMs, VLMs) will "act as a driver" in executing increasingly complicated driving decisions and tasks. Under this realm, a natural question arises: Can humans trust autonomous driving models? In our opinion, the key-point of trustworthy lies in the explainability of the autonomous driving models. They should be able to explain the reasons of making the decision to human drivers, not just execute driving actions. LLMs/VLMs are promising for enhancing trustworthy autonomous driving by providing high-level reasoning and understandable explanations in real-time.

\section{Conclusions} \label{sec5}
This survey provides the first systematic review of data-centric evolution in autonomous driving, including the big data system, data mining, and closed-loop technologies. In this survey, we start by formulating the dataset taxonomy categorized by milestone generations, reviewing the development of AD datasets throughout the historical timeline, introducing dataset acquisition, settings, and key features. Furthermore, we elaborate the closed-loop data-driven autonomous driving system from both academic and industrial perspectives. The workflow pipelines, procedures, and key technologies within data-centric closed-loop system are discussed in detail. Empirical studies are provided to demonstrate the utilization and advantages of data-centric closed-loop AD platforms for algorithm R\&D and OTA upgrading. Finally, a comprehensive discussion is carried regarding the Pros and Cons of existing data-driven autonomous driving technologies, as well as future research directions. Emphasis is placed on the new datasets beyond 3rd generation, hardware supports, personalized AD recommendation, explainable autonomous driving. We also express our concerns about Generative AI models, data security and trustworthy in the future development of autonomous driving.

\appendix


\section*{Acknowledgments}
We deeply appreciate Guangxin Su from Data \& Knowledge Research Group at University of New South Wales, Xingbo Du from ThinkLab at Shanghai Jiao Tong University, Yinfeng Xiang from State Key Lab of Industrial Control Technology at Zhejiang University for the discussion. We also sincerely thank dear colleagues from BYD Intelligent Driving R\&D Centre at Dept. of IDA in Shenzhen China, especially Shitai Zhang and Liang Zhao. Lincan Li have had a good time working with you together. 


\bibliographystyle{named}
\bibliography{ijcai24_abbrev_all}

\end{document}